# Image Segmentation Using Frequency Locking of Coupled Oscillators

Yan Fang[1*], Matthew J. Cotter[2+], Donald M. Chiarulli[3*], Steven P. Levitan[4*]

[*]University of Pittsburgh, [+]The Pennsylvania State University

[1]yaf13@pitt.edu, [2]mjc324@cse.psu.edu, [3]don@cs.pitt.edu, [4]levitan@pitt.edu

## ABSTRACT
Synchronization of coupled oscillators is observed at multiple levels of neural systems, and has been shown to play an important function in visual perception. We propose a computing system based on locally coupled oscillator networks for image segmentation. The system can serve as the preprocessing front-end of an image processing pipeline where the common frequencies of clusters of oscillators reflect the segmentation results. To demonstrate the feasibility of our design, the system is simulated and tested on a human face image dataset and its performance is compared with traditional intensity threshold based algorithms. Our system shows both better performance and higher noise tolerance than traditional methods.

## Categories and Subject Descriptors
C.1.3 [PROCESSOR ARCHITECTURES]: Other Architecture Styles - *Analog computers*, *Heterogeneous (hybrid) systems,* Neural nets

## 1. INTRODUCTION
Neural oscillation occurs at different scales in biological systems. Recently, theoretical studies on neuronal oscillation in visual perception and scene segmentation have made significant progress. In the study of synchronization of neural oscillations in cats' visual cortex, Eckhorn introduced a mammal neural model to emulate the mechanism of the visual cortex [1][2]. His work supported the theory that an object in a visual scene is represented by temporal correlations encoded by neural oscillations [3]. In this oscillatory correlation, each object is represented by a group of synchronized neural oscillators and different objects are represented by different groups that are not synchronized with each other [4].

Eckhorn's model was used in the Pulse Coupled Neural Network (PCNN) by Johnson [5], and soon was applied to the image segmentation problem [6]. However, Johnson's neuron model and network behavior being biologically inspired was quite complicated. To address this, Wang introduced a simple relaxation oscillator as a neuron model and proposed a Locally Excitatory Globally Inhibitory Oscillator Network (LEGION) [4][7][8].

VLSI implementations of LEGION for image segmentation have been performed [9][10][11][12]. But, the original LEGION model is limited to binary images. The stimuli to the oscillators can only be positive or negative. Later versions of the model use the coupling strength between nodes of the network to represent the intensity of grayscale image pixels [8]. However, this requires dynamically controllable coupling circuits, which makes it difficult to build large scale, high-speed, low power networks.

In this work we propose an oscillator network which uses frequency (rather than phase) locking, fixed nearest neighbor coupling, and simple oscillator models. We show that this model is suitable to a variety of oscillator models including neuronal, mechanical and chemical oscillators. The model performs as well as or better than traditional software algorithms for gray scale image segmentation and is more robust in the presence of noise.

We are motivated to look at a variety of oscillator models corresponding to both harmonic and relaxation oscillators due to the recent advances in emerging technologies for oscillatory based computation [13][14][15]. For many of these devices, we only have low level technology dependent models which are computationally expensive to simulate and subject to particular technology dependent parameters. Therefore, it would be good to know that our algorithms work for a wide range of possible models. Additionally, while much of that work focuses on clusters of fully connected oscillators, we are pursuing nearest neighbor networks which have different behaviors and are more suitable to computations which exhibit spatial locality.

The rest of this paper is organized as follows. First, we introduce three oscillator models and the local nearest neighbor network we use for our simulations. Then we consider the application of image segmentation as a representative task that can benefit from spatially locality. We perform several segmentation experiments showing the capabilities of the networks as compared to a standard segmentation algorithm. Finally we conclude with observations about the effectiveness and generality of these oscillator network models.

## 2. OSCILLATOR MODELS
In this section, we introduce three examples of oscillator models that have been developed and utilized for different applications. The first model is the neural oscillator in the LEGION model mentioned above [4]. The second is an oscillator model based on a chemical reaction called the Belousov-Zhabotinsky reaction [16]. The third one is an abstract model of a MEMS oscillator proposed by Hoppenstead and Izhikevich [17].

This work was funded, in part, by the National Science Foundation under the grants, INSPIRE Track 1: Sensing and Computing with Oscillating Chemical Reactions," DMR-1344178 and NSF "Expeditions in Computing Collaborative Research: Visual Cortex on Silicon," CCF-1317373.

*DAC'10*, Work-In-Progress, Jun 1–4, 2014, San Francisco, CA, USA.

## 2.1 Neural Oscillator Model

The basic building block of LEGION is a relaxation oscillator that is defined by the two differential equations:

$$\frac{dx_i}{dt} = 3x - x_i^3 - y_i + 2 + \rho + I_i + S_i \quad (1a)$$

$$\frac{dy_i}{dt} = \epsilon[\gamma(1 + \tan h\,(x_i/\beta)) - y_i] \quad (1b)$$

For oscillator (i), two variables $x_i$ and $y_i$ are respectively the excitatory unit and inhibitory unit. $\rho$ is a noise term introduced for desynchronizing between oscillators and $I_i$ represents the external simulation that controls the oscillation state. $S_i$ denotes the coupling term from other oscillators. $\rho$, $I_i$, and $S_i$ are three terms that can change the frequency of oscillation. The oscillator can be interpreted as a model of a spiking generator for a single neuron. If we do not consider the noise term and coupling term then when the external stimuli $I > 0$, the model becomes active and generates spikes periodically. When $I < 0$, the model receives inhibitory stimuli and does not oscillate. Figure 1 provides an example of the active and inactive states of oscillators with the nullclines, trajectory, and output waveform of $x$ and $y$. The x-nullcline ($dx/dt=0$) is a cubic curve and the y-nullcline ($dy/dt=0$) is a sigmoid curve. The intersection of two nullclines is the fixed point, whose position determines the state of oscillator. In the active state, when $I = +1$, the middle branch of x-nullcline is crossed by the y-nullcline as Figure 1(a) shows. The first derivatives of $x$ and $y$ along the trajectory form positive feedback paths to themselves, but inhibit each other, which generates the oscillatory behavior. While in the negative state, $I = -1$, and the left branch of the x-nullcline is crossed by the y-nullcline (Figure 1(b)) which generates a stable fixed point. Since $y$ will not change and $dx/dt = -x$ around the fixed point, any perturbation will be attracted back to this point, and no oscillation occurs. In this work, we only use the active mode. For $I > 0$, I controls the frequency of oscillation where higher stimuli lead to higher frequencies. This relaxation oscillator model can be easily implemented with circuits [10][11].

## 2.2 Chemical Oscillator Model

The second relaxation oscillator model we use in our system is the Belousov-Zhabotinsky (BZ) oscillator model. This model was implemented in analog CMOS circuits and fabricated to simulate the corresponding chemical reactions [16]. The BZ reaction is a periodic oxidation-reduction phenomenon in liquid-state reagents. It produces a variety of rhythms and orders in the form of propagating chemical waves [18].

An analog cellular-automation model for the BZ reaction is also proposed in [16] and the differential equations that describe the dynamics are:

$$\frac{d[x_1]}{dt} = \frac{1}{\tau}(-[x_1] + f([x_1] - [x_2], \beta_1)) + S \quad (2a)$$

$$\frac{d[x_2]}{dt} = -[x_2] + f([x_1] - \theta, \beta_2) \quad (2b)$$

Where $f(\cdot)$ is a sigmoid function defined by:

$$f(x, \beta) = \frac{1 + \tan h\,\beta x}{2} \quad (3)$$

Similar to the neural oscillator model in LEGION, the BZ oscillator has two variables $[x_1]$ and $[x_2]$, respectively to represent the concentration of $HBrO_2$ and Br- ions during the reaction. $\beta_1$ and $\beta_2$ are the parameters for two sigmoid functions. $\tau$ is the time constant that determines the frequency of oscillation. The BZ oscillator model also has two modes like the neural oscillator, here called the oscillation mode and the excitatory mode. In the first mode, the oscillator produces limit cycle oscillations, while it stays inactive and stable in the second mode. The state of BZ oscillator is also determined by the position of its fixed point, which is controlled by the value of $\theta$. Figure 2 shows the nullclines of the two variables and the waveform of $[x_1]$ in the two different modes. The detailed analysis of the dynamics is described in [16], which is very similar to the neural oscillators. In our work, we use the oscillation mode by configuring $\theta$.

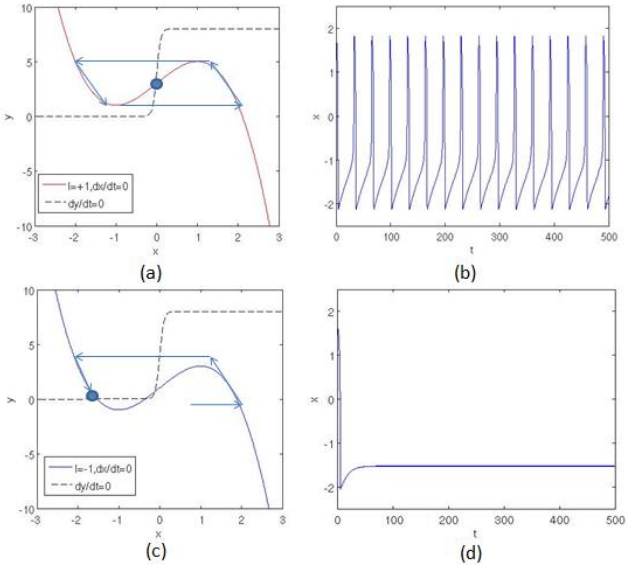

Figure 1 (a) Nullcline and trajectory of active state (I=1); (b) waveform of active state (I=1); (c) nullcline and trajectory of inactive state (I=-1); (d) single spike waveform of inactive state (I=-1). In these examples, ρ=0.02, e=0.1, γ=4, β=0.1, S=0

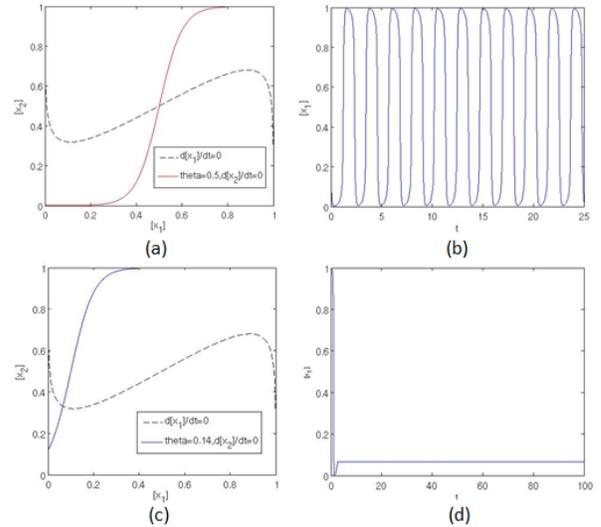

Figure 2 (a) Nullcline of oscillation mode ($\theta = 0.5$); (b) waveform of active state ($\theta = 0.5$); (c) Nullcline of excitatory state ($\theta = 0.1$); (d) Waveform of excitatory state ($\theta = 0.1$). In these examples, $\beta_1 = 5$, $\beta_2 = 10$, τ=0.06

## 2.3 Mechanical Oscillator Model

Besides the two relaxation oscillator models, we pick a MEMS oscillator model to test the compatibility of our system for a different style of oscillator. Unlike the spiking output of relaxation oscillators, this model behaves more like a traditional harmonic oscillator that generates a sinusoid signal. In Hoppenstead and Izhikevich's work, they model the behavior of a MEMS resonator and utilize this model to construct a Hopfield neural network for pattern recognition [17]. The details of the electro-mechanics can be found in [19].

The abstract mathematical model of a MEMS oscillator $i$ can be described by:

$$\frac{dz_i}{dt} = (c_i + i\omega_i)z_i + d_i z_i |z_i|^2 + S_i \quad (4)$$

Where $z_i$ is a complex variable, $c_i$ is a damping term, and $w_i$ is the natural frequency of the oscillator. $d_i$, denotes a nonlinear factor that ensures a stable amplitude and $S_i$ is the coupling term from the other oscillators. Figure 3 shows the limit cycle and waveform of the real part and the imaginary part of z. For this model, there is no excitatory or inactive state and the oscillator always oscillates.

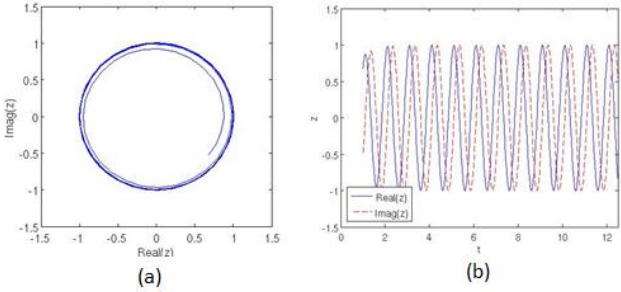

**Figure 3 (a) limit circle of oscillation; (b) The output waveform of z**

## 3. OSCILLATOR NETWORK

The structure of the oscillator network is a two dimensional array of coupled oscillators where each oscillator is coupled with its eight neighbors, shown in Figure 4. If we label the oscillators with their coordinates, the set of the neighbors $N_{ij}$ of oscillator $C_{ij}$ is defined by:

$$N_{ij} = \{C_{kl} | (|k - i| \leq r) \text{ and } (|l - j| \leq r)\} \quad (5)$$

Where $r$ is the neighbor radius, and $r = 1$. Thus, the coupling term $S_{ij}$ of each oscillator can be defined as:

$$S_{ij} = c \cdot \sum C_{kl} \in N_{ij} \quad (6)$$

where c is the coupling coefficient that represents the coupling strength between two oscillators.

In this oscillator network, coupling is bidirectional. Each oscillator computes the sum of output signals from its neighboring oscillators and also broadcasts its own output signal to them.

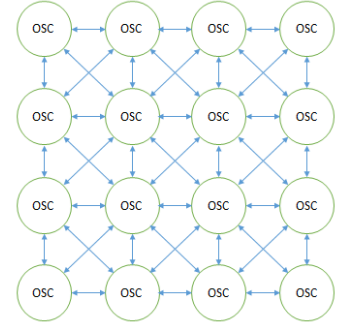

**Figure 4 Nearest Neighbor Network with bi-directional coupling**

For the image segmentation task, the oscillator network is configured such that it has the same size as the input image and each oscillator corresponds to one pixel. We initialize the oscillation of each oscillator with a frequency, depending on the intensity information of each pixel. In this work, we use grayscale images with the intensity ranging from 0 to 1. The actual control frequency parameter values for the oscillators are a mapping of pixel intensities so that the image is represented by a frequency band. For instance, for the neural oscillator we use the intensity of the input image to configure $I$ of each oscillator.

After initialization of the oscillator network, the oscillators try to synchronize with their neighbors and their frequencies begin to shift towards each other. If the pixels belong to the same region, usually their intensity values are close to each other. Thus the corresponding oscillators will synchronize with each other and lock to the same frequency. Otherwise, their frequencies are too far apart to synchronize and they keep oscillating with their own frequency. As a result, the oscillators in the network are clustered into groups. Oscillators within each group share the same or similar frequency, which differs from the other groups. Accordingly, pixels are clustered into different regions.

Once the network has converged, which means that oscillators' frequencies have become stable, we read out the array of each oscillator's frequency as the output and use the result in the final segmentation.

In an ideal case, the regions segmented from an image are labeled with different frequency values. However, some oscillators might fail to lock to the frequency of their own cluster. The reason could be that an oscillator represents a noise pixel or its location is on the boundary between regions. Under this situation, we can use the conventional intensity based segmentation techniques to process the histogram of frequency and to cluster pixels into regions, such as Otsu's method [20]. In this case, the oscillator network serves as a filter for pixel intensities. Therefore, even if the oscillators desynchronize with each other, their frequency difference can still provide information for the clustering of pixels. We show an example that illustrates this case in the Appendix Figures A1 and A2.

The shifting and locking of frequencies depends on three factors. The first factor is the initial difference of frequencies, which represents the intensity difference between pixels. Hence, if we map the pixels' intensities into wider frequency ranges, the difference between oscillators' frequencies will be enlarged and thus increase the difficulty of synchronization. The second factor is the coupling coefficient. High coupling coefficients mean stronger coupling strengths and fewer regions segmented as a result. Increasing the coupling coefficient can improve the noise resistance but may also deform the contours of the original image. The third factor is the spatial locality in the image, which comes from the geometric features in the image. Since each oscillator's neighboring oscillator is also coupled with other oscillators, its oscillation is actually influenced by oscillators further than its local neighborhood. This influence decreases as the distance increases and indirectly reflects the image gradient.

For different oscillator models, we configure the parameters and choose their frequency range in order to obtain optimal segmentation results. The values and ranges of parameters used for these experiments are:

- Neural oscillator model: $\rho = 0.02$, $e = 0.15$, $\gamma = 10$, $\beta = 0.1$, $I = 2 \text{ to } 4$, frequency range: 0.072 to 0.092.
- BZ oscillator model: $\beta_1 = 5$, $\beta_2 = 10$, $\theta = 0.5$, $\tau = 0.01\sim0.11$, frequency range: 0.32 to 0.56.
- MEMS oscillator model: $c = 1$, $d = -1$, $\omega = 2\pi\sim2.2\pi$, frequency range: 1 to 1.1.

Since we make no assumptions about the physical implementation of these oscillators and the simulations are done in Matlab, the oscillator frequencies are in arbitrary time units (AU).

## 4. IMAGE SEGMENTATION WITH OSCILLATORS

In investigating the quality of image segmentation methods we are faced with two problems. The first is to define what we mean by a good segmentation, and the second is to reasonably explore the parameters space for each of the oscillator methodologies.

Segmentation is typically one of the early steps of a complete image processing pipeline [21]. Therefore, "good" and "bad" are really defined by the subsequent stages. Given that the goal of segmentation is to identify regions of the image (pixels) that likely belong to the same object, we abstract out two measures of a noise-free segmentation. The first is to compare the segmented image to a human-generated ideal segmentation. The second is to compare the segmented image to "shape maximizing" segmentation. We discuss each of these below.

Our fundamental goal is not to prove that all or any particular oscillator network is better than any software algorithm for segmentation, but rather to show that simple, scalable, oscillator networks can perform segmentation on par with the state of the art software segmentation methods. Therefore, we have chosen only three of the myriad of oscillator models and a subset of all possible configuration parameters.

For the experiments we perform here, we use a dataset of 40 face images from the ATT Cambridge Image Database [22]. These are 32x32 pixel 256-level gray scale images, normalized to a range of 0-1. A subset of the images used, is shown in Figure 5(a). Figure 5(b) and (c) show two hand segmentations used as reference in the later experiments.

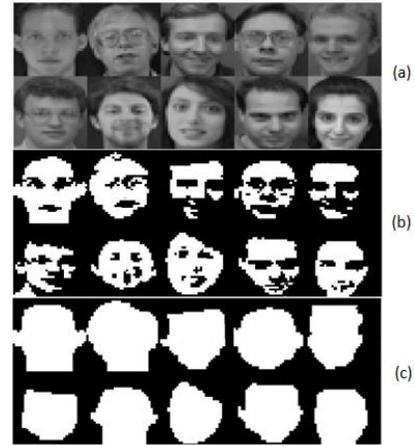

Figure 5: (a) 10 of the 40 face images used for testing; (b) reference detailed segmentation (c) reference maximum-region segmentation

### 4.1 Segmentation and Coupling

We first perform a simple test with one face image and a sweep of two important parameters: the degree of nearest neighbor coupling in the network and the threshold we use to separate the two frequencies corresponding to in/out of the primary segment. Figure 6 shows the result of these sweeps using the BZ model described above with a mapping of image intensity to $\tau$ values of 0.1 to 1.1 giving frequency ranges of $0.32 - 0.56$ (AU). We can see that, as the frequency threshold for clustering approached about 10% of the oscillator frequency all of the pixels were clustered into one "black" region, while for lower thresholds a high level of detail was observed. On the other hand, for clustering coefficient, $S$, above 0.3 a large region of pixels were put into a single cluster.

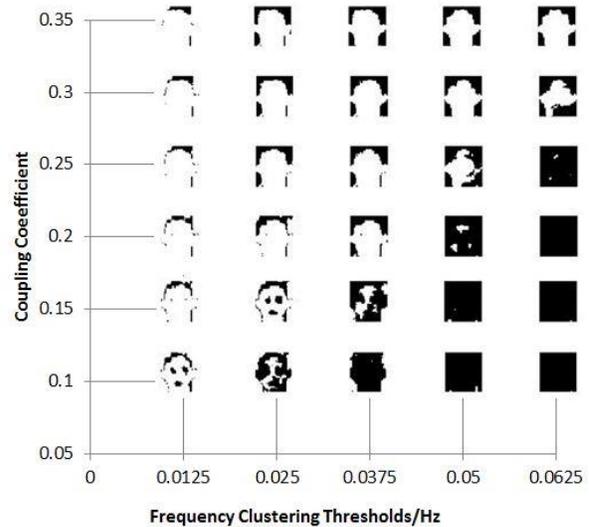

Figure 6: Relationship of coupling coefficient and frequency threshold to segmentation using the BZ oscillator network with $\tau$ = input image intensity, $\beta_1$ = 5, $\beta_2$ = 10 and $\theta$ = 0.5.

We can make two general observations about these results. First, either the figure in the lower left, or the one in the upper right could be considered a "good" segmentation of the image, depending on the needs of later image processing tasks, reinforcing a need for an abstract definition for comparison purposes. Second, is that the coupled oscillators perform clustering based on two criteria: spatial locality and intensity locality of the pixels in the image. It is the combination of these two (and their relative weight) which allows us to tune the networks to reduce noise sensitivity while maintaining good segment borders.

### 4.2 Comparison to Software Segmentation

To compare the three oscillator networks' performance to a software segmentation algorithm we chose an intensity based segmentation algorithm of Otsu's method [20]. This algorithm first bins the intensities into histograms and then segments the image based on clustering peaks in the histogram. The advantage of Otsu's method is that it optimizes the intensity threshold such that the sum of the standard deviations of the two histogram peaks is minimized. For the oscillator models we use coupling coefficients of $S_{BZ} = 0.1$, $S_{MEMS} = 0.05$, and $S_{Neural} = 0.02$ and the same thresholding scheme as Otsu, used on the oscillator frequencies. Using all 40 faces in the test set, the results are shown in Figure 7. This shows the percentage of mis-labeled pixels between the segmented image and the corresponding templates (in Figure 5(b)). We can see that the networks perform comparably to Otsu's method. We show that we can gain some improvement by tuning the threshold values in Appendix Figure A3.

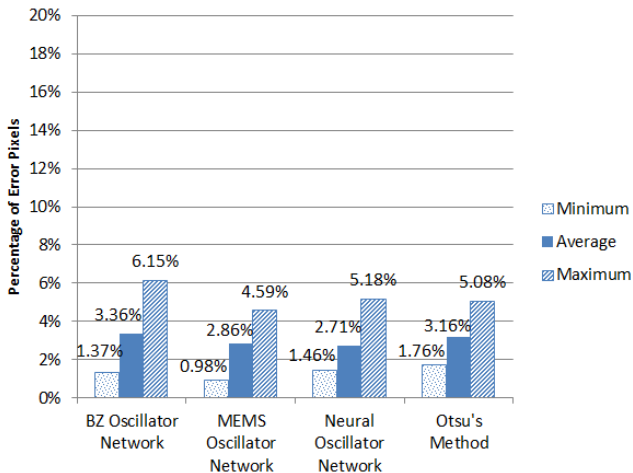

Figure 7: Comparison of segmentation performance using auto-scaled thresholds

### 4.3 Noise Sensitivity

To investigate the idea that spatial locality increases segmentation performance, we perform two experiments with additive Gaussian noise. While the oscillator networks use only local nearest neighbor coupling, the coupling is "transitive" in that coupled regions tend to grow and pull in adjacent pixel oscillators increasing the regions, and overriding noisy pixels. We can see this in the first line of Figure 8 where we show one face with increasing Gaussian noise with a variance from 0.005 to 0.03. The rest lines of Figure 8 provide the segmentation results from different algorithms.

Figure 9 gives the results for this test for each of the oscillator models and for Otsu's method. For these tests we defined an error as the number of pixels that were classified differently when compared to the classification that the same model did on the noise-free image. We can see that each of the oscillator models performs better than the locality oblivious software model.

Continuing to look at noise sensitivity we tested the methods on all 40 images adding Gaussian noise with a variance of 0.002. These results are shown in Figure 10, where we compare our results to the templates in Figure 5(c). Again all the network models do better than the software model.

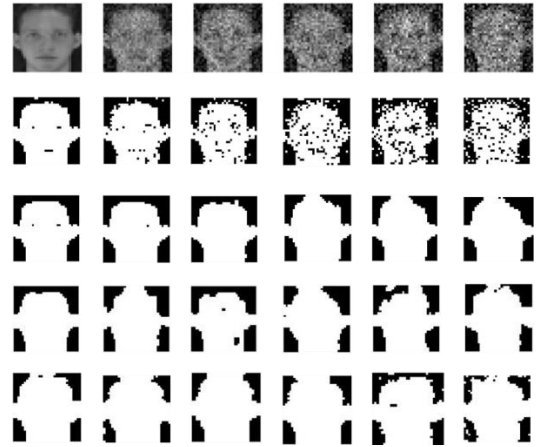

Figure 8: Illustration of segmentation results with increasing Gaussian noise

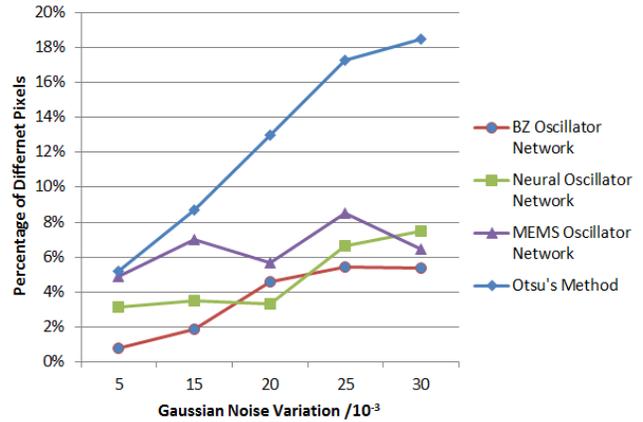

Figure 9: Segmentation performance with Gaussian noise

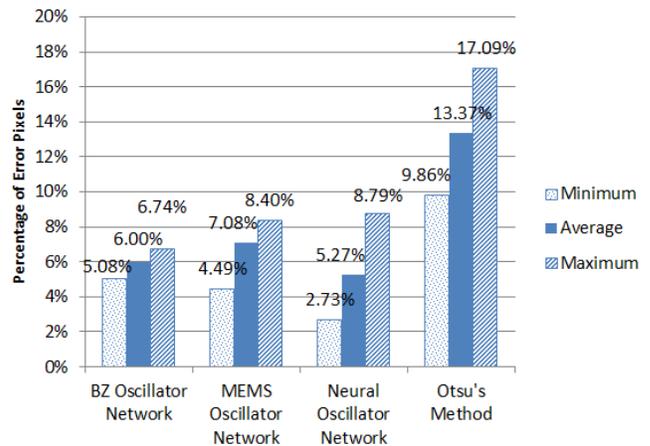

Figure 10: Segmentation performance on 40 noisy images

### 4.4 Segmentation for Shape

As we discussed earlier, sometimes the goal of segmentation is to find the largest contiguous shapes, rather than details of the image. In this case the ideal segmentations are closer to Figure 5(c). In this case we can again tune the parameters of the networks to optimize performance, increasing the coupling strengths to we use coupling coefficients of $S_{BZ} = 0.35$, $S_{MEMS} = 0.1$, and $S_{Neural} = 0.05$. The results are shown in Figure 11.

## 5. Conclusions

In this work we have shown the ability of networks of simple oscillators to perform "spatially local" computation. In particular we have shown that for image segmentation, oscillator arrays perform at least as well, if not better than state-of-the-art software techniques. Even for the cases where the oscillators do not completely "lock" the local coupling causes frequency shifting to create easily identifiable regions. The advantage of these systems is their ability to capture both spatial and intensity locality using large arrays of simple devices. We have also shown that both harmonic and relaxation based oscillator models work in this environment, leading us to conclude that these kinds of applications would be suitable to a wide variety of emerging nanoscale oscillator technologies with the potential of low power and high speed computing.

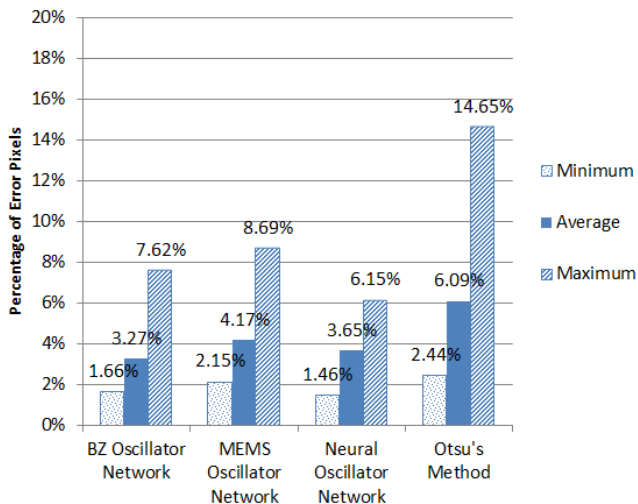

**Figure 11: Performance for shape extraction**

## 6. ACKNOWLEDGMENTS

The authors would like to thank Vijaykrishnan Narayanan, Gert Cauwenberghs, Suman Datta, Philip Wong, and the rest of "Visual Cortex on Silicon" research group.

# Appendix

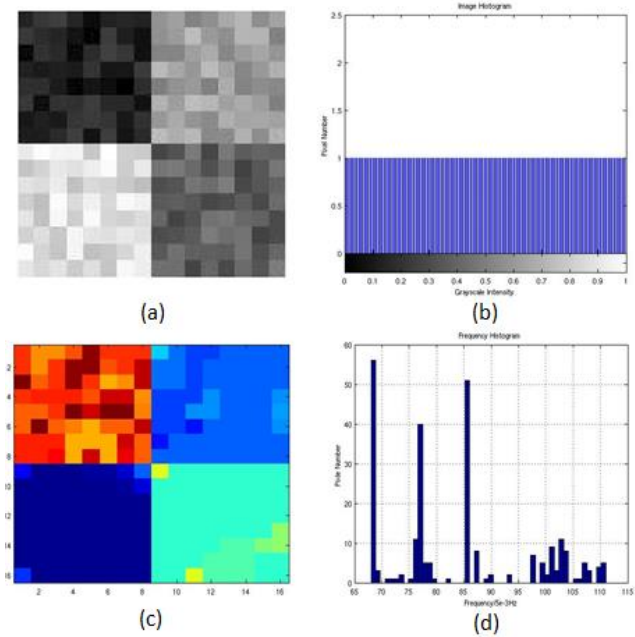

Figure A1. An example of how oscillator networks can segment regions based on the geometric features of image. Since the intensity of image is uniformly distributed, it contains no information for intensity based segmentation, the oscillators are still grouped into clusters though frequency locking. From the histogram of output frequencies, we can notice that some oscillators do not synchronize. It is helpful to use a thresholding technique on the frequency histogram. Figure A1(a) A 16 by 16 8-bit grayscale image that consists of 4 square regions with different average intensities; (b) The intensity histogram of this image, a uniform distribution; (c) The output frequency matrix from the oscillator networks, represented by a color spectrum; (d) The histogram of output frequencies. In this case we use the BZ oscillator model with the parameter set in Section 3.

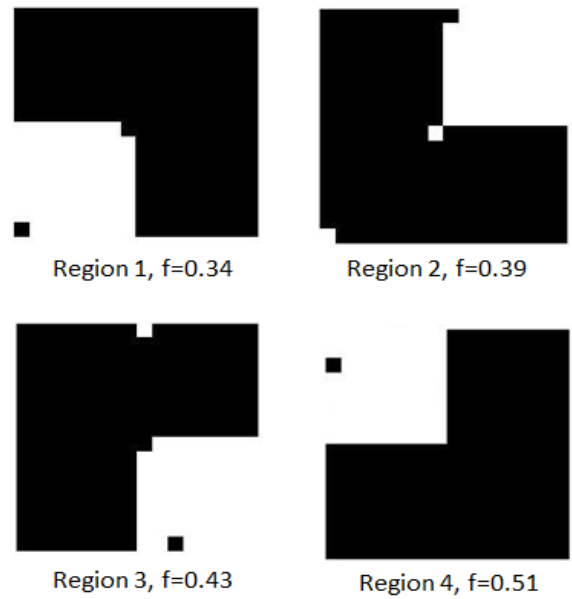

Figure A2. Segmentation results obtained from Figure A1(d). The frequency threshold is 0.025 used to segment regions 1,2,3. Since region 4 has more desynchronized oscillators, the threshold needed to be 0.15 for this result.

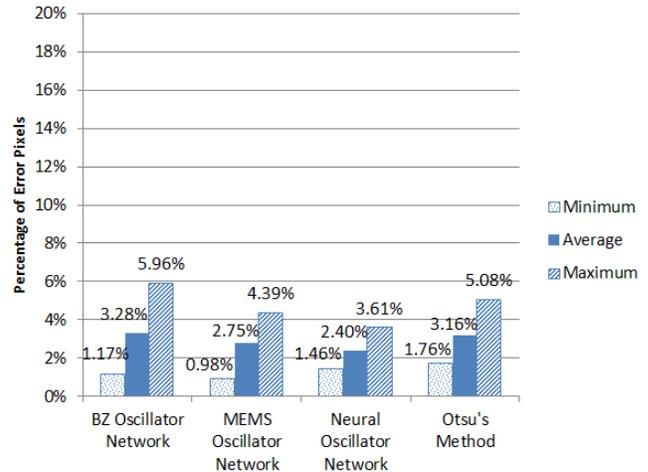

Figure A3: Comparison of segmentation performance using tuned thresholds. This result shows the same test with modifications in the threshold, T, to optimize the networks' performance. $T_{BZ} = 0.0125$, $T_{MEMS} = 0.02$, $T_{Neural} = 0.025$. We can see some improvement over the previous case which we hypothesize comes from the locality information that is not available in the software model.